\documentclass[letterpaper, 10 pt, conference]{ieeeconf}
\IEEEoverridecommandlockouts

\usepackage{cite}
\usepackage{amsmath,amssymb,amsfonts}
\usepackage{algorithmic}
\usepackage{graphicx}
\usepackage{textcomp}
\usepackage{xcolor}
\def\BibTeX{{\rm B\kern-.05em{\sc i\kern-.025em b}\kern-.08em
    T\kern-.1667em\lower.7ex\hbox{E}\kern-.125emX}}

\usepackage[bookmarks=true,breaklinks]{hyperref}

\usepackage{algorithm}

\usepackage{balance} 

\usepackage{mathtools}

\graphicspath{{figures/}}

\usepackage[caption=false,font=footnotesize]{subfig}

\usepackage{array}

\begin{document}

\title{\LARGE \bf
Reinforcement Learning for Reduced-order Models of Legged Robots
}

\author{Yu-Ming Chen$^{1}$, Hien Bui$^{1}$ and Michael Posa$^{1}$   \thanks{$^{1}$The authors are with the General Robotics, Automation, Sensing and Perception (GRASP) Laboratory, University of Pennsylvania, Philadelphia, PA 19104, USA
        {\tt\small \{yminchen, xuanhien, posa\}@seas.upenn.edu}}
        }

\maketitle

\setcounter{footnote}{1}

\begin{abstract}

Model-based approaches for planning and control for bipedal locomotion have a long history of success. 
It can provide stability and safety guarantees while being effective in accomplishing many locomotion tasks.
Model-free reinforcement learning, on the other hand, has gained much popularity in recent years due to computational advancements. 
It can achieve high performance in specific tasks, but it lacks physical interpretability and flexibility in re-purposing the policy for a different set of tasks. 
For instance, we can initially train a neural network (NN) policy using velocity commands as inputs. However, to handle new task commands like desired hand or footstep locations at a desired walking velocity, we must retrain a new NN policy.
In this work, we attempt to bridge the gap between these two bodies of work on a bipedal platform.
We formulate a model-based reinforcement learning problem to learn a reduced-order model (ROM) within a model predictive control (MPC). 
Results show a 49\% improvement in viable task region size and a 21\% reduction in motor torque cost.
All videos and code are available at \href{https://sites.google.com/view/ymchen/research/rl-for-roms}{https://sites.google.com/view/ymchen/research/rl-for-roms}.

\end{abstract}

\section{Introduction}\label{sec:RL_intro}

\begin{figure*}[t!]
 \centering
 \includegraphics[width=1\linewidth]{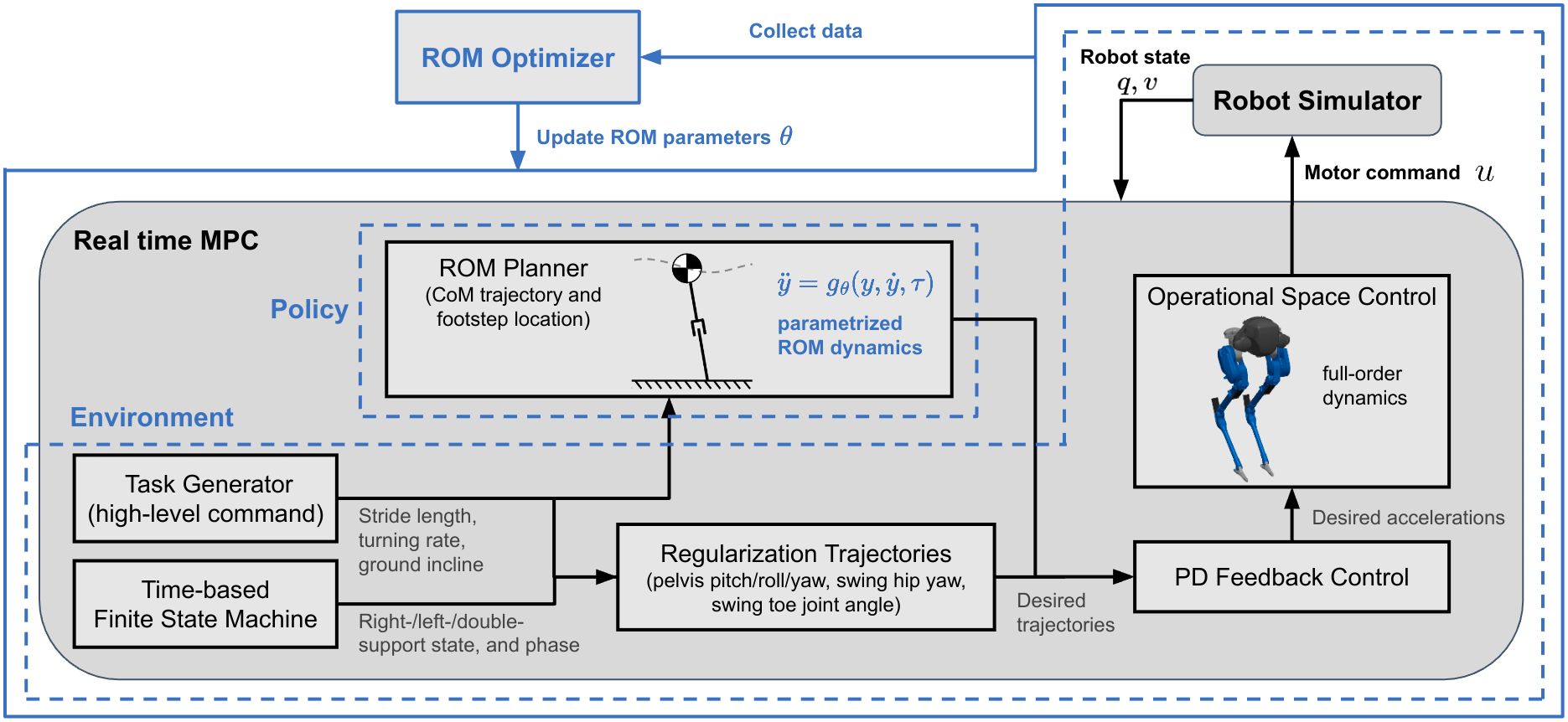}

 \caption{
 The diagram of our reinforcement learning framework for reduced-order models (ROMs) of legged locomotion.
 We learn a ROM in simulation where the robot is controlled via a real time model predictive control (MPC).
 The MPC follows a high-level command (task) and operates based on a time-based finite state machine (FSM), governing footstep timing in left-support, right-support, or double-support state.
  The MPC contains two trajectory generators -- a reduced-order model planner and a regularization trajectory generator to fill out the joint redundancy of the robot.
  All desired trajectories are converted to desired acceleration command via PD feedback control before being sent to the Operational Space Controller (OSC), a quadratic-programming-based inverse-dynamics controller \cite{sentis2005control, wensing2013generation}. 
 In this learning framework, the ROM planner is the policy, while everything else in the closed-loop system, such as the simulation and OSC, constitutes the environment. 
 We parameterize the policy using ROM parameters, specifically the parameters of ROM dynamics.  
 The optimizer collects data from simulation rollouts and updates the model (policy) parameters according to a user-specified reward function.
 To ensure that the optimizer collects data at a consistent rate, we maintain a fixed update rate of 20Hz for the ROM planner, and we downsample the environment state (which operates at 1000Hz) to match this 20Hz rate.
 }
 
 \label{fig:rl_and_mpc_diagram}
\end{figure*}

Model-based planning and control have shown significant success for many years in navigating legged robots\cite{Kuindersma14, ackerman2012boston, AtlasMPC, gibson2022terrain, xiong20223, wensing2022optimization}.
For example, Boston Dynamics's bipedal robots performed parkour and natural walking \cite{AtlasMPC, PETMANPrototype}, 
while SRI's humanoid DURUS achieved high energy efficiency in walking \cite{reher2016realizing, ackerman2015durus}.
Moreover, having robot models makes it possible to analyze the system stability and provide stability or safety guarantees via techniques such as capturability \cite{koolen2012capturability}, sums of squares \cite{posa2017balancing}, hybrid zero dynamics \cite{westervelt2003hybrid} or control barrier functions \cite{ames2019control, nguyen20163d, khazoom2022humanoid}.  
However, as these methods can be computationally costly, reduced-order models (ROMs) are frequently deployed to achieve real time planning and control, albeit at the cost of reduced performance \cite{wensing2022optimization, chen2023beyond}.

On the other hand, model-free reinforcement learning (RL) has emerged as a powerful tool for automatically synthesizing high-performance control policies \cite{siekmann2021blind, crowley2023optimizing, dao2022sim, ma2023learning, miki2022learning}. 
Siekmann et al. highlighted the robustness achieved by a neural network (NN) policy in blind stair-walking \cite{siekmann2021blind}, 
while Ma et al. developed a policy utilizing a robot's arm for fall damage mitigation and recovery \cite{ma2023learning}.
These works demonstrate that neural networks, as universal approximators, enable robots to excel in specific tasks. 
However, model-free polices lack interpretability, rendering the existing model-based stability and safety techniques unsuitable. 
Additionally, the model-free methods struggle with generalizing policies to new task parameters without policy retraining, as task space parameterization is determined during the training phase. 
For example, the policy described in \cite{ma2023learning} effectively recovers a fallen robot but does not consider walking velocity commands. 
To enable the robot to walk, it is necessary to incorporate these velocity commands as additional inputs to the NN and undergo a fresh policy training process.
Moreover, robots may encounter a wide array of potential tasks, such as footstep timing, adaptive limb adjustment during hardware failure, head movement for item detection, whole-body interactions between arms and legs, or collaborative furniture carrying with humans.
Thus, it is impractical to enumerate all possible tasks to avoid future policy retraining.

This paper attempts to combine model-based planning and control with reinforcement learning to obtain the best of both worlds in the context of bipedal locomotion.
It aims to retain the physical intepretabilty and the task flexibility of the model-based approach, while utilizing the RL capability in maximizing robot performance.
Specifically, we use model predictive control (MPC) as the model-based control policy and learn a model within the MPC that maximizes the robot's performance via RL.

Within the MPC, ROMs are often used in order to achieve real time planning\cite{chignoli2021humanoid, xiong2020dynamic, gibson2022terrain}.
Classical approaches to reduced-order modeling often seek to find low-dimensional models which approximate some collected data (e.g. in approximating the solutions to fluid dynamics\cite{peherstorfer2015dynamic}). 
For a controlled system, such as a bipedal robot, the data is necessarily a product of the control policy, and thus the data and the ROM-induced controller are invariably intertwined. 
Nonetheless, recent work in locomotion has attempted this classical approach via model-based RL \cite{yang2020data}, 
or via robust control (treating the gap between a ROM and the full model as a disturbance \cite{pandala2022robust}).

In contrast, we jointly optimize over the ROM and the induced control policy. Robot actuators are partially utilized to ensure that the system behaves like the reduced-order model, up to the limits of control performance. With this perspective, a good ROM is not one that matches some existing set of robot behaviors. Rather, we find task-optimal ROMs that maximize robot performance when deployed in conjuction with modern model-based planning and control architectures.
Our prior work \cite{chen2023beyond} optimizes controller-agnostic ROMs via trajectory optimization.
However, the performance of the ROM may not seamlessly transfer to the robot due to the influence of specific heuristics or designs of the controller.
In this paper, we improve our prior work by taking into account the control policy used on the robot while optimizing the ROM.
This effectively closes the performance gap between offline model optimization and online model deployment.

\subsection{Contributions}

\begin{enumerate}
\item Formulate a reinforcement learning problem to directly optimize, with respect to a user-specified objective function, a reduced-order model given a controller.
\item Demonstrate that the RL algorithm improves the model performance by evaluating the model performance in detail, compared to both an existing model (LIP) and an optimal model from our prior work. 
\item Demonstrate the flexibility of our model-based approach in task space parameterization in the post-training phase.
\end{enumerate}

\vspace{-1mm}

\section{Background}\label{sec:background}

\subsection{Full-order Model of Cassie}
Cassie \cite{ackerman2017agility} is a two-legged robot with closed-loop linkages and compliant components (leaf springs).
It has line-feet, so there is one degree of freedom of underactuation when it stands on one foot.
In the full dynamics, we model the linkages using (holonomic) distant constraints, and we account for the reflected inertia of the motors and also joint frictions.
Let $q\in \mathbb{R}^{23}$ and $u\in \mathbb{R}^{10}$ be the configuration and input of the full-order model respectively. 
Let $v$ be the velocity of the robot's configuration, and $x=[q,v]$ be the state of the robot.

\subsection{ROMs for Legged Locomotion}\label{sec:ROM_short_intro}

Let $y$ and $\tau$ be the configuration and input of the reduced-order model of a robot respectively.
The reduced-order model can be defined by two functions -- an embedding function and the dynamics of the ROM 
\begin{subequations}\label{eq:model}
\begin{align}
y&=r(q),  \label{eq:model_kin}\\
\ddot{y}&= g(y,\dot{y},\tau). \label{eq:model_dyn}
\end{align}
\end{subequations}
Reduced-order models for legged locomotion commonly describe the center of mass (CoM) motion, 
while each model imposes different constraints on the robot to derive its own CoM dynamics \cite{Kajita91, blickhan1989spring}. 
For example, the linear inverted pendulum model (LIP) \cite{Kajita91, Kajita01} assumes zero vertical CoM accelerations, and the spring-loaded inverted pendulum model (SLIP) \cite{blickhan1989spring} enforces a spring-mass dynamics.
Inspired by this, we search for an optimal CoM dynamics $g$, while using the same embedding function $r$, the CoM position relative to the stance foot, as LIP and SLIP.
In this case, $\dim (y) = 3$. 
Furthermore, we assume $\dim (\tau) = 0$ for simplicity, although we keep $\tau$ in Eq. \eqref{eq:model}, since ROMs can have inputs in general (e.g. angle torques or the center of pressure \cite{Kajita91}).

\subsection{Model Predictive Control}\label{sec:RL_mpc}

Model predictive control (MPC) is a control technique that plans for desired input trajectories by predicting the robot's state in a receding horizon fashion \cite{chignoli2021humanoid, xiong2020dynamic, gibson2022terrain}. 
For legged robots \cite{ackerman2017agility, NadiaHumanoidthesis, AtlasHumanoidthesis}, the full models are usually too high dimensional to allow real time planning, so common ROMs introduced in Section \ref{sec:ROM_short_intro} are used instead. 
Our MPC for Cassie is outlined in Fig. \ref{fig:rl_and_mpc_diagram}. 
It consists of a few components -- a high-level command generator, a finite state machine, trajectory generators (including a ROM planner) and a low-level controller.

\subsubsection{High-level Command Generator}
At the beginning of the diagram is a high-level command generator (a task generator). 
It outputs the desired stride length, the desired turning rate and the perceived ground incline level.

\subsubsection{Finite State Machine}
The MPC follows a predefined step timing dictated by the finite state machine.
This determines the ground contact sequence inside the ROM planner and the contact mode inside the Operational Space Control.

\subsubsection{Trajectory Generator}
The primary trajectory generator creates a plan for the desired CoM state and desired footstep location given a high-level command.
Specifically, the ROM planner solves an reduced-order trajectory optimization problem which minimizes an objective function while satisfying a set of constraints.
The objective function predominantly minimizes the tracking error for the high-level commands.
The constraints account for the initial state of the ROM, the dynamics of the ROM, and the input and state constraints.
In this case, the inputs are the footstep locations, so the input constraints take care of the feasible stepping location.
For more details of the planner, we refer to Section IV of \cite{chen2023beyond}.
Besides the planner, we also need to specify desired trajectories for the remaining degree of freedom of the robot. We generate these regularization trajectories by simple heuristics such as maintaining a horizontal attitude of the pelvis body, having the swing foot parallel to the contact surface, and aligning the hip yaw angle to the desired heading angle.

\subsubsection{Low-level Controller}
All desired trajectories (functions of time) from the trajectory generators are passed into PD feedback controllers where they are converted into desired accelerations. 
Finally, Operational Space Control is an inverse dynamics controller formulated as a quadratic programming problem \cite{sentis2005control, wensing2013generation}. 
It takes into account the full model dynamics 
in the optimization problem and outputs an optimal input $u$ that minimizes the tracking error of the desired accelerations.

\section{Problem Statement}\label{sec:RL_high_level_statement}

Our goal is to find the optimal model parameters that maximize the performance of the robot given a task distribution $\Gamma$. 
In this study, $\Gamma$ is a uniform distribution over a range of stride length, turning rate and ground incline.
Let $u(\theta)$ be a model-based control policy (i.e. a controller) that guides the robot to follow the trajectories of a ROM parameterized by $\theta$.
From another viewpoint, this same policy $u(\theta)$ also constrains the robot to behave like the ROM.
Let $\{\left(x_t,u_t\right) \mid t = 1, 2, ..., T\}$ be the state and input trajectories of the robot completing a task $\gamma \sim \Gamma$ under the policy $u(\theta)$.  We evaluate the performance for this task $\gamma$ and policy $u(\theta)$ using a cost function $h(x,u)$ accumulated over the trajectories: \begin{equation}\label{eq:accumulated_cost}
H_{\gamma,u(\theta)} =  \displaystyle\sum_{t=1}^T h\left(x_t,u_{t}\right). 
\end{equation}
Given this evaluation metric, an optimization problem for finding the model can be formulated as 
\begin{equation}\label{eq:high_level_problem_openloop}
\underset{\theta}{\text{min}} \  
\mathbb{E}_{\gamma \sim \Gamma} \ 
\underset{u(\theta)}{\text{min}} 
	\left[ H_{\gamma,u(\theta)} \right]
,  
\end{equation}
where the inner-level optimization minimizes the cost $H_{\gamma,u(\theta)}$ over all possible controllers $u(\theta)$ that constrain the robot to behave like a ROM parameterized by $\theta$,
and the outer-level optimization minimizes the expectation of inner-level cost over a task distribution.
Eq. \eqref{eq:high_level_problem_openloop} has been proven to be solvable by our prior work \cite{chen2023beyond} with successful results.
However, the optimal policy $u(\theta)$ of \eqref{eq:high_level_problem_openloop} is not computable online in real time, necessitating an alternative policy $u_{o}(\theta)$ for online model deployment. 
Thus, while solving \eqref{eq:high_level_problem_openloop} leads to a ROM capable of maximal performance, this performance is not necessarily realizable via real time control, leading to reduced closed-loop performance.
To improve this, in this paper, we find the model parameters $\theta$ while using the control policy $u_{o}(\theta)$ during offline training: 
\begin{equation}\label{eq:high_level_problem_closedloop}
\underset{\theta}{\text{min}} \  
\mathbb{E}_{\gamma \sim \Gamma} 
	\left[ H_{\gamma,u_{o}(\theta)} \right]
.  
\end{equation}
Specifically, this control policy $u_{o}(\theta)$ is the model predictive control presented in Section \ref{sec:RL_mpc}.
This seemingly simple change necessitates a significant shift in algorithmic approach, which we detail in Section \ref{sec:learning}.

\section{ROM learning via RL}\label{sec:learning}

In this section, we cast the model optimization problem in Eq. \eqref{eq:high_level_problem_closedloop} as a model-based reinforcement learning problem.

\subsection{Reinforcement Learning Structure}

In reinforcement learning, the system comprises a policy and an environment.
The policy takes the current environment state $s$ and outputs an action $a$. 
Given the current state $s$ and the action $a$, the environment transitions to the next state $s'$. 
Each pair of state and action $(s,a)$ results in a reward $r$. 
In this work, the policy is the ROM planner, 
and the environment includes everything else in the closed-loop system (Fig. \ref{fig:rl_and_mpc_diagram}).
Given this choice, the state of the environment $s$ includes the robot's state and input $(x,u)$, task $\gamma$, and finite state machine information (including the phase of the current state). 
The policy action $a$ are the CoM states and the foot step locations (the solution of the ROM trajectory optimization).
Additionally, we limit the policy's update rate to 20Hz, so that the state and action pairs are collected at a fixed rate.

\subsection{Policy Parameterization}

We parameterize the ROM with monomials of the state of the ROM, with linear weights. That is, 
\begin{equation}
\ddot{y}= g_\theta(y,\dot{y},\tau) = \Theta \phi(y,\dot{y}), \label{eq:model_dyn_parametrized}
\end{equation}
where $g_\theta$ is the ROM dynamics parameterized by $\theta$, $\Theta \in \mathbb{R}^{n_y \times n_\phi}$ is the matrix form of $\theta$, and $\phi$ is the feature vector containing the monomials.
Even though we use monomials here, any function approximator (e.g. a neural network) can be used to parameterize the ROM dynamics in Eq. \eqref{eq:model_dyn_parametrized}.
We note that the ROM parameters $\theta$ are the policy parameters, as the ROM planner is the policy in our RL framework (Fig. \ref{fig:rl_and_mpc_diagram}).

In this paper, we initialize the ROM to an LIP, because the LIP is effective for simple walking tasks, and this initialization speeds up the learning process.
To implement this initialization, we augment the feature vector $\phi$ with the terms in the LIP dynamics function.
Additionally, we use monomials of order up to 2. That is, $\phi$ includes elements such as $y_1$, $y_0^2$ and $y_0 \dot{y}_1$, where the subscripts denote the indices of each element in $y$.
Given this, $\theta$ is of dimension 90.
The parameterization choice in Eq. \eqref{eq:model_dyn_parametrized} preserves physical interpretability, since the learned model describes the CoM dynamics.

\subsection{Rewards and the RL Problem Statement}\label{sec:reward}

Instead of minimizing the cost in Eq. \eqref{eq:high_level_problem_closedloop}, our RL formulation maximizes the return (i.e. accumulated rewards) $R=\sum_{t=1}^T r_t$,
where $r_t$ is the reward at time $t$.
Therefore, to encourage minimizing the cost $h(x,u)$ and achieving a desired task $\gamma$,
we design the reward function \begin{equation}\label{eq:total_reward}
r = \text{exp}(-w\cdot h) + 0.5 \ \text{exp}\left(-\|\gamma - \gamma_{fb}\|_W\right), 
\end{equation}
where $w$ and $W$ are constant weights, $\gamma$ is the desired task value, and $\gamma_{fb}$ is the achieved task value (a function of the robot's state).
We note that this reward is a function of the environment state $s$.
Furthermore, the accumulated reward structure already incentivizes the robot to finish the entire episode, eliminating the need for penalty terms in case of early simulation termination (e.g. the robot falls).
In this paper, we choose $h=u^T u$ (quadratic cost on motor torques), 
and the horizon $T=100$ which is equivalent to 5 seconds of simulation time.
Additionally, the user-specified $h$ is ultimately the metric for model performance evaluation.

The objective of our RL problem is to maximize the expected return over the task distribution:
\begin{equation}\label{eq:RL_high_level_objective_two_level}
\underset{\theta}{\text{max}} \  
\mathbb{E}_{\gamma \sim \Gamma} \left[ R \right]
.
\end{equation}
There are several algorithms for solving Eq. \eqref{eq:RL_high_level_objective_two_level}, such as policy gradient and evolutionary strategy.  
In this work\footnote{
Proximal Policy Gradient (PPO) also resulted in successful model training in our experiment, but it required more hyperparameter tuning. 
}, we choose Covariance Matrix Adaptation Evolutionary Strategy (CMA-ES)  \cite{hansen2006cma}.
The advantage of this approach includes easy parallelization \cite{salimans2017evolution} and easy hyperparameter tuning (partially due to the absence of value approximation).
We use the package Optuna \cite{optuna_2019} for the CMA-ES optimizer.

Let $p_\theta$ be the probability distribution over $\theta$ in the CMA-ES algorithm.
The exact problem that CMA-ES solves is
\begin{equation}\label{eq:RL_high_level_objective_two_level_CMA}
\underset{p_\theta}{\text{max}} \  
\mathbb{E}_{\theta \sim p_\theta} \left[
\mathbb{E}_{\gamma \sim \Gamma} \left[ R \right]
\right].
\end{equation}
The difference between Eq. \eqref{eq:RL_high_level_objective_two_level_CMA} and \eqref{eq:high_level_problem_closedloop}, besides the cost-reward difference, is the stochasticity of parameters $\theta$ needed for the exploration of CMA-ES.

\begin{algorithm}[!t]
\caption{Evaluation of return $\hat{R}\left(\theta, \{\gamma_j \}\right)$}
\begin{algorithmic}[1]\label{alg:return_eval}
\renewcommand{\algorithmicrequire}{\textbf{Input:}}
\renewcommand{\algorithmicensure}{\textbf{Output:}}
\REQUIRE  model parameters $\theta$ and sampled tasks $\{\gamma_j\}$ 
\FOR {$j=1,\ldots,N_\gamma$}
\STATE Roll out an episode with the MPC in Section \ref{sec:RL_mpc}
\STATE Compute the return $R_{\gamma_j} = \sum_{t=1}^T r_t$ of this rollout
\ENDFOR
\RETURN $\frac{1}{N_\gamma} \sum_{j=1}^{N_\gamma} R_{\gamma_j}$
\end{algorithmic}
\end{algorithm}

\begin{algorithm}[!t]
\caption{CMA-ES with curriculum learning}
\begin{algorithmic}[1]\label{alg:cma}
\renewcommand{\algorithmicrequire}{\textbf{Input:}}
\renewcommand{\algorithmicensure}{\textbf{Output:}}
\REQUIRE  initial mean $\theta_0$ and variance $\sigma_0^2$ for the parameter distribution $p_\theta$, and initial task set $\Gamma_d$
\FOR {$\text{iter}=1,2,\ldots$}
\IF{mod(iter, $N_c$) = 0}
\STATE Grow the task set $\Gamma_d$ (Section \ref{sec:curriculum_learning})
\ENDIF
\STATE Randomly draw $N_\gamma$ tasks $\{\gamma_j \}$ from $\Gamma_d$
\STATE Sample $N_\theta$ parameters $\{\theta_i \} \sim p_\theta$
\FOR {each sampled $\theta_i$}
\STATE Compute return $\hat{R}\left(\theta_i, \{\gamma_j \}\right)$
\ENDFOR
\STATE Update the mean and covariance of $p_\theta$ (CMA-ES)
\ENDFOR
\end{algorithmic}
\end{algorithm}

\subsection{Curriculum Learning}\label{sec:curriculum_learning}

We observe that learning a policy for a large task space is difficult,
so we implement a curriculum learning similar to \cite{margolis2022rapid} to facilitate the learning process.
Our method discretizes the continuous task domain of $\Gamma$ into a set $\Gamma_d$.
This set $\Gamma_d$ starts small and expands every $N_c$ iterations by including the adjacent tasks of successful tasks during learning.

To evaluate the inner expected value in Eq. \eqref{eq:RL_high_level_objective_two_level_CMA}, we approximate it by randomly drawing $N_\gamma$ number of tasks from the set $\Gamma_d$ and averaging the returns of these tasks.
Let $\hat{R}$ be this approximate expected return given model parameters $\theta$. 
The evaluation of $\hat{R}$ is shown in Algorithm \ref{alg:return_eval}.
As a part of the curriculum learning, we also adjust the number of tasks $N_\gamma$, as the size of $\Gamma_d$ grows larger.
Specifically,
$N_\gamma = \text{floor}\left(\rho_\gamma  |\Gamma_d|\right),$
where $\rho_\gamma$ is the sampling percentage and $|\cdot|$ counts the number of elements in a set.

The algorithm for solving Eq. \eqref{eq:RL_high_level_objective_two_level_CMA} with curriculum learning is outlined in Algorithm \ref{alg:cma}. 
In every iteration, CMA-ES samples a few model parameters $\theta$'s according to $p_\theta$, and evaluates the value (return $\hat{R}$ in this case) for each sampled $\theta$. Given these values, CMA-ES updates the mean and the covariance matrix of the parameter distribution $p_\theta$.

\section{Experimental Result}\label{sec:RL_experiment}

\begin{figure*}[!t]
\centering

\includegraphics[width=0.7\linewidth]{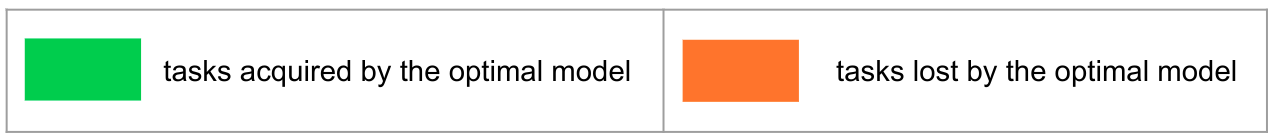}

\subfloat[Comparison against LIP (initial model)]{
\includegraphics[width=0.47\linewidth]{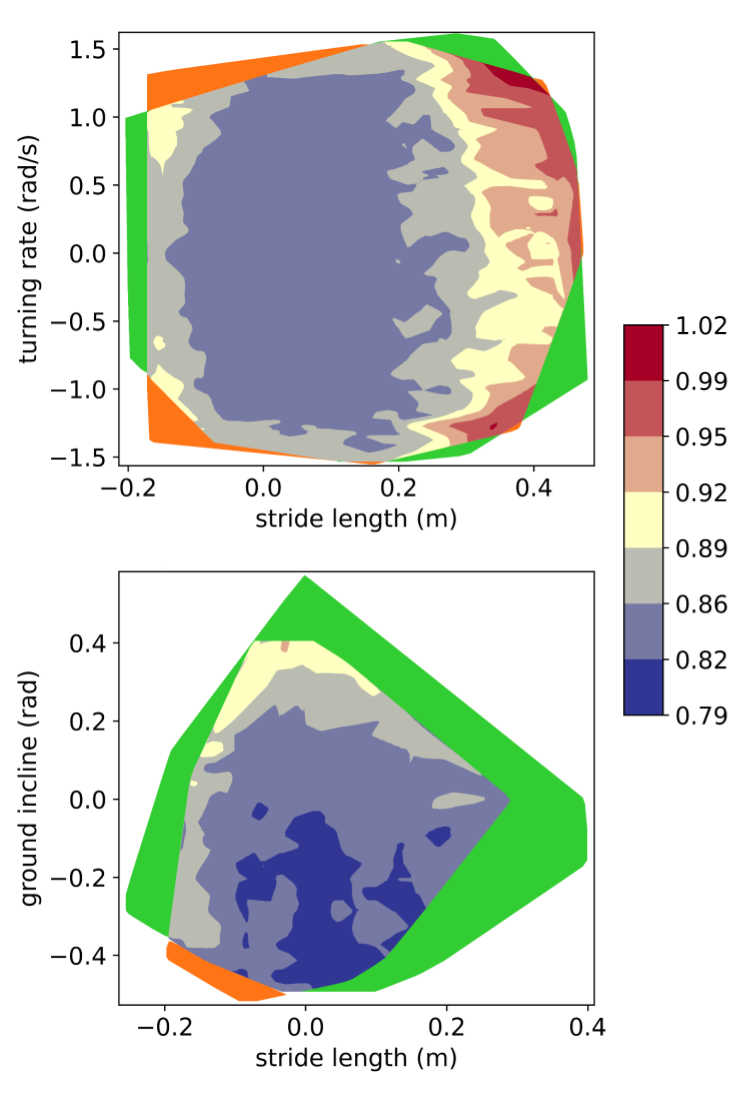}
\label{fig:cost_landscape__springed_cassie__RL}}
\hfil
\subfloat[Comparison against the optimal model of prior work's approach]{
\includegraphics[width=0.47\linewidth]{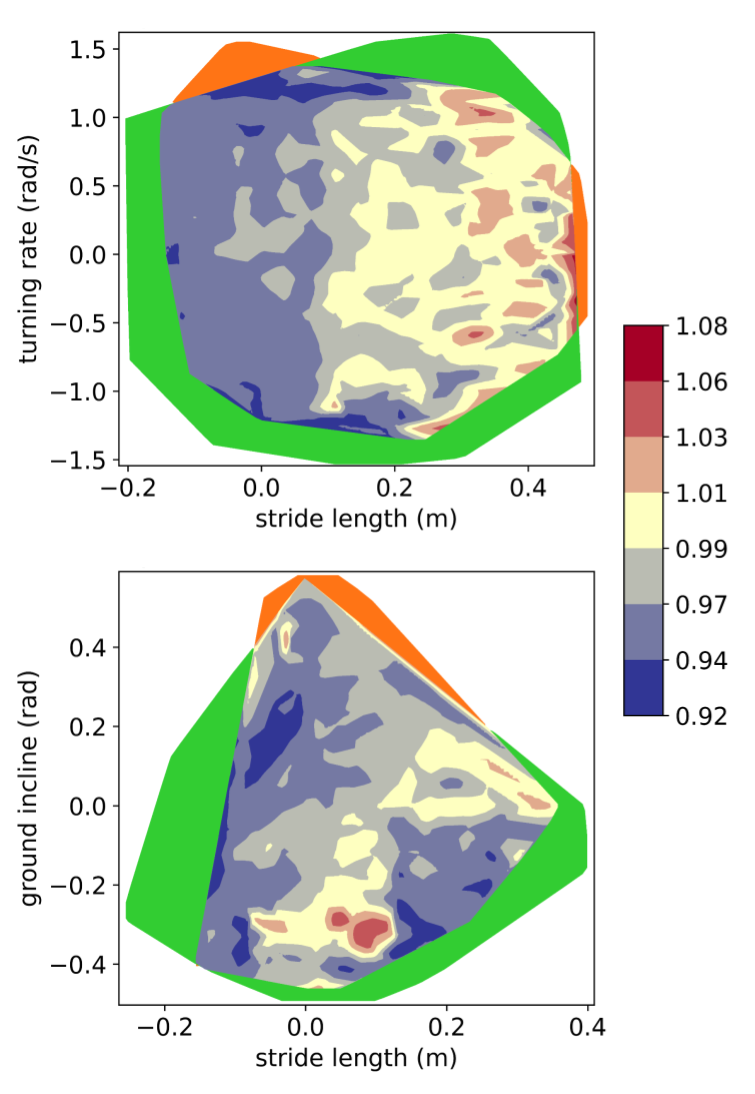}
\label{fig:cost_landscape__springed_cassie__RL_vs_TRO}}
		
\caption{
Cost landscape comparisons. 
Fig. \ref{fig:cost_landscape__springed_cassie__RL} compares the optimal model to the initial model.  
Fig. \ref{fig:cost_landscape__springed_cassie__RL_vs_TRO} compares the optimal models between this paper and the prior work.  
For each model comparison, we create landscapes in two sets of experiments -- one involving turning rate and stride length (with a constant ground incline of 0 rad), and the other involving ground incline and stride length (with a constant turning rate of 0 rad/s).
Each plot shows the ratio of the optimal model's cost to the compared model's cost. 
The color scheme red-to-blue illustrates the degree to which the optimal ROM shows improvement.
Ratio below 1 means the optimal ROM performs better than the compared model, and vice versa. Green color shows the task regions gained by the optimal model, and orange color shows the task regions lost by the optimal model.
}
\label{fig:cost_landscape_comparison__springedCassie}
\end{figure*}

We learn a ROM in simulation using Drake \cite{drake2019} offline, and deploy it to the same simulation environment for detailed evaluation comparison between the initial model and the optimal model.
We also compare this optimal model to the one derived using the prior approach \cite{chen2023beyond}.
Lastly, we showcase the flexibility in task space reparameterization in Section \ref{sec:new_task_space_parameterization}.

\subsection{Hyperparameters}

The hyperparameters for the RL are shown in Table \ref{table:hyperparam}. 
The number of parameters sampled per iteration $N_\theta$ is adaptively chosen by the CMA-ES optimizer according to the parameter dimension.
The sample density $\rho_\gamma$ is chosen to be 0.1 which has the benefit of speeding up the learning process compared to evaluating all discretized task samples (i.e. $\rho_\gamma=1$). 
The initial standard deviation $\sigma_0$ of the parameters is set to a relatively small number, since the initial ROM already works for simple tasks. 
Moreover, we observe that using a larger standard deviation $\sigma_0$ does not show better performance in our experiments, and it has a drawback of potentially diverging from the optimal parameters from time to time.
For the curriculum learning, we set the initial task space to be straight-line walking with stride lengths between -0.1 m and 0.2 m.
The task space is discretized by 0.1 m, 0.1 rad and 0.45 rad/s in stride length, ground incline and turning rate, respectively.
Lastly, we only expand the task space every 30 iterations to learn a model gradually.

\subsection{Comparing the Optimal Model to LIP (Initial Model)}\label{sec:compared_to_lip}

We compare the performance of the initial model $\theta_0$ and the optimal model from the learning result.
To evaluate the model performance, we run the simulation for a wide range of stride lengths, ground inclines and turning rates.
We then extract the periodic walking gaits according to the criteria shown in Table \ref{table:periodic_criteria} which ensure variations over foot steps fall below specific thresholds.
Given these periodic trajectories, we compute the cost $H_{\gamma,u(\theta)}$ and plot the landscape of the cost ratio of the optimal model to the initial model, shown in Fig. \ref{fig:cost_landscape__springed_cassie__RL}.
The color scheme red-to-blue illustrates the degree to which the ROM shows improvement, with red corresponding to a minimal improvement and blue to a 20\% cost reduction.
In addition to the ratio, we also visualize the task region where either the initial model or the optimal model fails to complete, visualized by green and orange.
Green corresponds to the task regions that the optimal model gains, and orange corresponds to the regions that the optimal model loses.

In this example, the optimal ROM reduces the cost across almost the entire task space (up to 21\% cost reduction). 
For flat ground walking tasks, the optimal model shows the largest improvement in the region of small stride lengths.
For inclined walking, the optimal model achieves higher performance improvement in downhill tasks than uphill tasks.
Additionally, the optimal model increases the task region size by 49\% for inclined walking.
For example, at an incline of -0.2 rad, the maximum stride length increases from 0.2 m to 0.37m.

\subsection{Comparison against our Prior Work}

We also conducted a set of experiments similar to Section \ref{sec:compared_to_lip}.
Instead of comparing against the initial model, we compare the optimal model from Algorithm \ref{alg:cma} to the optimal from the prior work \cite{chen2023beyond}.
The cost landscape comparison for this is shown in Fig. \ref{fig:cost_landscape__springed_cassie__RL_vs_TRO}.

Compared to the prior approach, the optimal model of this paper shows up to 28\% increase in task space along with an average 2.7\% cost improvement.
For the flat ground walking tasks, the model yields a 22\% larger viable task region than that of the prior approach.
For the inclined walking tasks, it is 28\% larger.
We hypothesize that this improvement comes from the fact that in the RL framework we are able to use a Cassie model of high fidelity during training, while the prior approach was limited to a simplified Cassie model for ease of solving the inner-level optimization in Eq. \eqref{eq:high_level_problem_openloop}.

We note that, besides the above numerical comparisons, the RL approach in this paper is easier to implement than the prior approach, since it does not require a careful implementation of offline trajectory optimization (inner-level optimization in Eq. \eqref{eq:high_level_problem_openloop}) in conjunction with the online control policy ($u_o(\theta)$ in Section \ref{sec:RL_high_level_statement}) for reducing the gap between the open-loop and closed-loop performances.

\subsection{Generalization to New Task Space Parameterization}\label{sec:new_task_space_parameterization}

One advantage of using a model-based approach is the flexibility in specifying new tasks, as motivated in Section \ref{sec:RL_intro}.
During the training stage of our experiment shown above, we train the model using common tasks including stride length, turning rate and ground incline.
We demonstrate that the model can be easily extended to achieve unseen tasks by modifying the MPC diagram in Fig. \ref{fig:rl_and_mpc_diagram}.
For example, we might want to ensure that a body-mounted sensor is oriented at a target of interest, or we might want a robot to collaboratively carry a table with a human, which requires the robot facing a different direction than the walking direction.
To mimic these scenarios for Cassie, we turn Cassie's pelvis to the side while walking forward.
We achieve this by simply changing the desired value of the pelvis yaw angle for the regularization trajectory generator in Fig. \ref{fig:rl_and_mpc_diagram}. 
In the accompanying video, we can see that the robot achieves this task without any more offline training.

\begin{table}[t!]
\centering
\begin{tabular}{ | c | c |  } 
 \hline
 initial standard deviation $\sigma_0$ & 1e-3 \\ 
 \hline
 number of sampled parameters per iteration $N_\theta$  & 17  \\ 
 \hline
 sample density $\rho_\gamma$  & 0.1  \\ 
 \hline
 number of iterations for task space expansion $N_c$  & 30  \\ 
 \hline
 stride length discretization & 0.1 m  \\ 
 \hline
 turning rate discretization & 0.45 rad/s  \\ 
 \hline
 ground incline discretization & 0.1 rad  \\ 
 \hline
\end{tabular}

\vspace{1mm}
\caption{Hyperparameters for the model learning}

\label{table:hyperparam}
\end{table}

\begin{table}[t!]
\centering
\begin{tabular}{ | c | c | } 
 \hline
 Stride length variation & $<2$ cm \\ 
 \hline
 Side stepping variation & $<3$ cm \\ 
 \hline
 Pelvis height variation & $<3$ cm \\ 
 \hline
 Pelvis yaw variation & $<0.1$ rad \\ 
 \hline
 Window size & 4 consecutive footsteps \\ 
 \hline
\end{tabular}

\vspace{1mm}
\caption{Criteria to qualify a periodic walking gait
}
\label{table:periodic_criteria}
\end{table}

\section{Conclusion and Future Work}\label{sec:rl_conclusion}

We formulate a model-based reinforcement learning problem for reduced-order models of legged locomotion.
This provides an avenue to bridge the gap between the well-established model-based control and the   emerging field of reinforcement learning for legged robots,
combining the performance-maximizing capability of RL with the physical interpretability and the task specification flexibility of model-based approaches.
The experiments show that the optimal model reduces the torque cost by up to 21\% and improves the viable task region size by up to 49\% over the traditional models like LIP.
We also compare this work to our prior work which uses full model trajectory optimization during the ROM optimization, 
and the results show an up to 28\% improvement in the viable task region size along with a mild improvement in torque cost.

In this paper, we solve the RL problem using CMA-ES, 
but theoretically any reinforcement learning optimizer can be applied to our RL framework in Fig. \ref{fig:rl_and_mpc_diagram}, 
since our policy (ROM planner) is differentiable \cite{amos2018differentiable}.
Future work includes exploring more optimizers to find one that results in the highest performance improvement. 
Additionally, we observed in our prior work \cite{chen2023beyond} that parametrizing the embedding function $r$ along with the ROM dynamics function $g$ (i.e. both Eq. \eqref{eq:model_kin} and \eqref{eq:model_dyn}) lead to higher performance improvement than parameterizing only $g$.
In this paper, we simplified the problem and the RL formulation by limiting $r$ to a CoM kinematic function, 
so one future direction is to parameterize both $r$ and $g$, and learn the model in a similar pipeline.

\section{Acknowledgment}

Toyota Research Institute provided funds to support this work.

\bibliographystyle{ieeetr}
\bibliography{library,yuming_library}

\balance

\end{document}